# Trust as indicator of robot functional and social acceptance. An experimental study on user conformation to iCub answer

Ilaria Gaudiello[1], Elisabetta Zibetti[1], Sébastien Lefort[2], Mohamed Chetouani[3], Serena Ivaldi[4]

igaudiello@gmail.com; ezibetti@univ-paris8.fr, sebastien.lefort@lip6.fr, mohamed.chetouani@upmc.fr ; serena.ivaldi@inria.fr

1. Cognition Humaine & Artificielle (CHART EA 4004), Univ.Paris 8, and EPHE Paris; LUTIN Living Lab - FED 4246 (Cité des Sciences et de l'Industrie, Paris). F-93526 Saint Denis - Cedex 02, France.

2. LIP6, Sorbonne Univ. UPMC Univ Paris 06 CNRS, UMR 7606, F-75005 Paris, France

3. Institut des Systèmes Intelligents et de Robotique (ISIR), CNRS & Sorbonne Univ., UPMC Univ. Paris 06, UMR7222, F-75005 Paris, France

4. INRIA, Villers-les-Nancy, F-54600 ; Loria, CNRS & Univ. de Lorraine, Loria, UMR 7503, Vandoeuvre-lès-Nancy, F-54500, France

**Abstract** To investigate the functional and social acceptance of a humanoid robot, we carried out an experimental study with 56 adult participants and the iCub robot. Trust in the robot has been considered as a main indicator of acceptance in decision-making tasks characterized by perceptual uncertainty (*e.g.*, evaluating the weight of two objects) and socio-cognitive uncertainty (*e.g.*, evaluating which is the most suitable item in a specific context), and measured by the participants' conformation to the iCub's answers to specific questions. In particular, we were interested in understanding whether specific (i) user-related features (i.e. desire for control), (ii) robot-related features (*i.e.*, attitude towards social influence of robots), and (iii) context-related features (*i.e.*, collaborative vs. competitive scenario), may influence their trust towards the iCub robot. We found that participants conformed more to the iCub's answers when their decisions were about functional issues than when they were about social issues. Moreover, the few participants conforming to the iCub's answers for social issues also conformed less for functional issues. Trust in the robot's functional savvy does not thus seem to be a pre-requisite for trust in its social savvy. Finally, desire for control, attitude towards social influence of robots and type of interaction scenario did not influence the trust in iCub.

Results are discussed with relation to methodology of HRI research.

**Introduction**

Developing efficient human-robot partnership demands acceptance of robots in our daily routines and environment. Acceptance is a sensitive topic of investigation in HRI research (*e.g.*, Fridin & Belokopytov, 2014; de Graaf & Allouch, 2013; Young, Hawking, Sharlin, *et al.* 2009; Heerink, Kröse, Wielinga, *et al.*, 2009; 2010; Salvini, Laschi, & Dario, 2010; Welch, Lahiri, Warren, *et al.*, 2010; Weiss, Bernhaupt, Lankes, *et al.*, 2009; Riek & Robinson, 2008; Kaplan, 2004). Traditionally defined as "the demonstrable willingness within a user group to employ information technology for the tasks it is designed to support" within the field of the User Acceptance of Information Technology (Dillon, 2001), acceptance takes on a new significance when referred to robots.

Through an attentive survey of the literature, we can identify six connotations of robot acceptance: representational, physical, behavioral, functional, social, and cultural[1] (see Gaudiello, 2015 – Phd dissertation). Among these dimensions, functional and social acceptances appear to play a key-role for effective human-robot interaction (Fridin *et al.*, 2014; Weiss *et al.,* 2009; Picard & Daily, 2005; Zaad & Allouch, 2008).

Functional acceptance refers to the level of perceived ease of use, usefulness (Heerink *et al.*, 2009; Weiss *et al.*, 2009), accuracy (Beer, Prakash, Mitzner, *et al.*, 2011; Schaefer, 2013), and innovativeness (Rogers, 1995; Kaplan, 2005) of the robot. Social acceptance covers a variety of issues such as social presence (Riether, 2013; Hamill & Harper, 2006), perceived sociability – often overlapping with social abilities and social intelligence (Heerink, 2010),

---

[1] *Representational acceptance* describes the influence of the mental models (stereotypes from science fiction, representation borrowed from preexistent mental models of technology, animals or children, etc.) on robot acceptance in terms of place attributed to the robot within common sense of ontology and suitable functions or roles of a robot. *Physical acceptance* defines the embodiment features of the robot - its morphology (anthropomorphic, zoomorphic, machine-like), size (human or animal size vs. household appliance or toy size, etc.) materials (*e.g.,* organic vs. mechanic), etc – that could impact robot acceptance in terms of likeability and credibility of the robot. *Behavioral acceptance* concerns those behaviors – for example proxemics (*i.e.,* intimate vs. public distance, direction of approach, etc.), and communication (*e.g.,* verbal, gestural, etc.) – that can affect acceptance in terms of believability and fluency of interaction. Finally, *cultural acceptance* refers to fundamental issues of a given culture, for example appetite for technology, social care, educational values, etc. which might alter acceptance in terms of intention to use and duration of use.

and social influence (Nomura, Suzuki, Kanda, *et al*., 2006) of robots.

Interestingly, while representational (Gaudiello, Lefort & Zibetti, 2015) and cultural (Li, Rau & Li, 2010) acceptance continuously vary with relation to user progressive familiarization to robots (Bartneck, Suzuki, Kanda, *et al*., 2007), and while we are likely to accept that a robot might have a rough physical aspect (Turkle, 2011), or to excuse the robot for an inappropriate behavior like we would do with a child (Young *et al*., 2009), we are much less likely to accept a useless robot (Kaplan, 2005) and we can be deceived by a robot that lacks social abilities (Shaw-Garlock, 2009; Heerink, 2010).

Today, robots endowed with functional and social abilities constitute two distinct categories on the market (Shaw-Garlock, 2009), with distinct purposes and labels - *e.g.*, 'utilitarian' and 'hedonistic' robots (Lee, Shin & Sundar, 2011). But robotic industry is definitely moving towards robots which are both functional and social (Shaw-Garlock, 2009) and researchers have advanced that users behavior will evolve accordingly: the more efficient a robot is, the more people would be likely to trust them on all levels, including social level (Young *et al*., 2009).

Indeed, contemporary research points out that, when it comes to real interaction situations, functional and social acceptance share a fundamental common ingredient: *users' trust in the robot* (Hancock, Billings, Schaefer, *et al.*, 2011; Shinozawa, Naya, Yamato, *et al.*, 2005). Trust is thus increasingly employed to assess the quality of human-robot interaction (Hoffman, Lee, Woods, *et al.*, 2009; Schaefer, 2013; Lee & See, 2004; Yagoda, 2011; van den Brule, Dotsch, Bijlstra, *et al.*, 2014; Kaniarasu, Steinfeld, Desai, *et al.*, 2012) and, as it emerges from this literature, it constitutes a valid indicator of robot functional and social acceptance (Yagoda, 2011).

For example, suppose a person is asking its domestic robot to help in deciding which basket of tomatoes is the heaviest. In this case the person would have to trust the robot functional answer (functional acceptance). Now, suppose the same person is asking to the robot which is the necessary equipment to put in the swimming pool bag of his child (a swimming cap? Or maybe water sandals?). In this case the person would have to trust in the robot social answer (social acceptance). The question is: will the person trust the robot's answer in both these cases? As showed in our example, functional and social acceptance are characterized by the same dynamics, but apply to very different issues, such as technical estimation *vs.* human care, and presuppose very different built-in knowledge and performance of the robot.

However, so far, the research models developed in the last decade to assess functional acceptance of robots (Heerink *et al*., 2009; Weiss *et al*., 2009; Fridin *et al.*, 2014; Beer *et al*., 2011) mainly borrow their conceptual frame from older technology acceptance models (Davis, Bagozzi & Warshaw,1989; Rogers, 1995; Venkatesh, Morris, Davi, *et al*., 2003) thus neglecting the specificity of the robot as a social agent. Several studies however have validated a significant set of indicators of social acceptance such as performance expectancy, attitude towards technology, anxiety, etc. (Weiss *et al*., 2009; see Heerink, 2010 for a review of those indicators).

But while those indicators can be useful to estimate the users' willingness to accept robots in a early phase of the interaction, trust can be a more eloquent indicator for daily and sustained interactions, in which users rely on robots to take decisions about functional and social issues. To explain our position, let us go back to the tomatoes and swimming pool example.

This kind of daily activities (*i.e.*, exchange information, take decisions, affirm one's point of view, accept compromise), can be considered as problems (Popper, 1991). To solve these problems we need to retrieve and apply knowledge that relates to either scientific rational thinking or natural thinking - *i.e.*, to common sense. These two kinds of thinking are based on different process, finalities, and content (Piaget, 1967; Guimelli, 1999; Jack, Dawson, Begany, *et al*., 2013). Scientific thinking seeks the truth through logical demonstration, and it is supposed to result in a true solution. Here the decision-making process relies on the examination of the objective knowledge on the physical world (Piaget, 1967; Popper, 1991). The decision thus responds to the physical world itself and it is validated by the application of the rules (in our example, these rules are represented by conventional measures of weight). On the opposite, natural thinking or common-sense (beliefs, knowledge, opinions) works according to a subjective and/or social logic, whose finality is the achievement of pragmatic solution, that is of the more appropriate solution in a given context (Guimelli, 1999). Here, the decision-making process relies on subjective knowledge, and social norms (*e.g.,* Abric, 1994a, b). The decision thus responds to personal evidence, which is connected to the common-sense knowledge acquired through everyday experience.

Now, in order to accept the robot help with relation to the first problem (the weight of the tomatoes basket) the person needs to rely on the robot's scientific knowledge and technical ability (*e.g.*, its force sensors) to provide an objective judgment. On the contrary, in the second case (the choice between the head-cup and the thongs) the person needs to rely on the robot common-sense knowledge. So, in both cases there is an underlying inferential process by which the person bases his/her trust on the robot knowledge or "savvy" (Young *et al.*, 2009) called into the decisional process.

Functional savvy refers thus to robot's ability to efficiently operate by assuring useful and accurate performances with relation to the functions it was designed for (Weiss *et al.*, 2009; Fridin *et al.*, 2014; Heerink, 2010). The tacit assumption about functional savvy is that, just like computers, the robot is equipped with scientific knowledge about the physical world and with powerful instruments (*e.g.*, sensors) to make much more precise measurements than a human can do. This ability is considered a necessary condition by users to decide to employ a robot (Kaplan, 2005; Heerink *et al.*, 2009; Beer *et al.*, 2011).

On the opposite, social savvy describes the robot's capability to fit into the social structures and activities of a given context (Young *et al.*, 2009) according to its role in the interaction (Welch *et al.*, 2010). The tacit assumption about functional savvy is that the robot possesses common-sense knowledge, that is, knowledge of situations, behavioral scripts and current norms that are in use (Dauthenhan, 2007; Hamill *et al.*, 2006).

Therefore, within the limited scope of this research, we define functional acceptance as users trust on robot functional savvy, and social acceptance as users trust on robot social savvy.

However, caution should be used, since, as mentioned social savvy requires an adaptive context-dependent knowledge, which cannot be hardwired into the robot and demand complex learning algorithms which are continuously improved (Lockerd & Brazeal, 2004). Consequently, even if users do have expectations about robot social savvy (Lohse, 2010; Coeckelbergh, 2012), at the present stage of robots development, these expectations are rarely confirmed (Duffy, 2003; Fong, Nourbakhsh & Dautenhahn, 2003).

This explains why researchers presently call for the necessity of building measures which could assess not only trust on robot functional savvy but also fine psychological aspects of non-obvious trust on robot social savvy (Heerink, 2010).

Several subjective measures of trust in the HRI research field have been developed and are mostly based on self-report (*i.e.*, questionnaires). If these measures allow to collect users overt opinions they also tend to induce a reflective mental posture, and are then limited in their capacity to register those spontaneous opinions and inner beliefs that could allow to better understand on which robot knowledge the users base their trust and which are the possible relations that they establish between the functional and the social savvy.

An alternative and complementary approach to test trust in machines comes from the classical experimental paradigm of the Media Equation Theory (Nass & Moon, 2000; Nass, Fogg & Moon, 1996; Nass, Moon, Fogg & Reeves, 1995). According to this theory, when engaged in collaborative tasks, people tend to unconsciously accept computers as social entities: they trust on answers provided by the computer and they conform to its answers. Nass and his colleagues introduce then the concept of *conformation* to investigate users spontaneous or 'mindless' response to computer as an objective measure to evaluate levels of trust in machine and its acceptance as a potential partner. Likewise, one of the research challenge today is thus to design an experimental paradigm that enables to register those mindless reactions towards robots which are susceptible of revealing users trust on robot, with particular attention for trust on its functional and/or social savvy.

For this reason, we have set the focus of our inquiry on the observation of users trust behaviors as an indicator of functional and social acceptance. Furthermore, inspired by the work of Nass and his colleagues, we have proposed to adapt the well known Media Equation Theory[2] paradigm by employing users' conformation to robots decisions as a new objective measure of human-robot trust during a human–humanoid decision making tasks under uncertainty (Yu, 2015).

By confronting users to tasks where they have to verbally express their decisions about functional and social issues, we assess whether the fact of experiencing perceptual uncertainty (*e.g.*, evaluating the weight of two slightly different objects) and socio-cognitive uncertainty (*e.g.*, evaluating which is the most suitable item in a specific context) leads users to withdraw their own decision and conform to the robot decision.

Moreover, relying on those studies where it was proven that trust can generally vary according to individual and contextual differences (Hancock *et al.*, 2011), we are also interested in identifying a set of factors which are likely to correlate with trust in robot functional and social savvy – such as individuals propensity to control (Burger & Cooper, 1979), fear of being influenced (Nomura *et al.*, 2006) and the context (competitive *vs.* collaborative) in which the decision making task involving the person and the robot take place.

---

[2] The Media Equation paradigm was transposed from human-computer interaction to human-robot interaction in a study by Kidd (2003). However, our research intention is to adapt this paradigm to the specificity of the robot as a functional and social technology rather than to transpose it: instead of being confronted with a unique task such as the Desert Survival Problem, our participants are confronted with two tasks, respectively functional and social.

Hence, the main research questions we address in the present work are:

(i) do users trust in robot functional and social savvy?
(ii) is trust in functional savvy a pre-requisite for trust in social savvy?
(iii) which individual and contextual factors are likely to influence this trust?

In order to answer to those questions we carried out an experimental study based on a human-robot interaction with fifty-six adults and the humanoid robot iCub. In the following, we will further illustrate the main issues of our research and bring evidences to justify the experimental design choices of our investigation with regard to pre-existing studies in HRI research.

## 1. Trust as a fundamental indicator of acceptance

Trust can determine the overall acceptance of a system (Parasuraman & Riley, 1997). For this reason, there is a considerable number of studies seeking a better understanding of the human-robot trust dynamics (see Hancock Billings, Schaefer, *et al.*, 2011 for a meta-analysis).

Broadly speaking, trust on system automation is classically defined as having confidence in the system to do the appropriate action (Biros, Daly & Gunsch, 2004) with personal integrity and reliability (Heerink *et al.*, 2009). Further definitions of trust borrowed from studies on human-human trust, commonly known as interpersonal trust (Mayer, Davis, & Schoorman, 1995; Rotter, 1971), focusing on expectations, common goals, uncertainty, and reliance as core elements of trust (Billings, Schaefer, Lloren *et al.,* 2012). In this sense, trust describes the expectation that a robot will help achieve an individual's goals in a situation characterized by uncertainty and by reliance of the individual on the robot (Hoffman *et al.,* 2009; Lee *et al.*, 2004).

The empirical works on human-robot trust carried in the latter half of the last decade (*e.g.*, Lee *et al.*, 2004; Yagoda, 2011; Burke, Sims, Lazzara, *et al.*, 2007) employ a variety of measures that aim at detecting changes in users level of trust and also factors that might decrease or enhance it (Schaefer, 2013), as we will outline in the following chapters in order to argue the interest to consider conformation as an objective measure of trust in the robot functional and social savvy.

**1.1 Commonly used measures of human-robot trust**

Two main categories of methods for assessing human-robot trust emerge in the literature: one is based on subjective or explicit measures and another on objective or implicit measures.

Objective measures can be retrieved from behavioral data (*e.g.*, response time) unconsciously produced by individuals (Hofmann, Gawronski, Gschwendner, *et al.*, 2005), whereas subjective measures (*i.e.*, questionnaires, self-report) can be retrieved from collected verbal data (*e.g.*, opinions) consciously produced by the individuals. If the former are limitedly developed in HRI, the latter are widely used.

There are very few studies on the relation between trust and behavioral cues. The most notable study in the field of HRI is from DeSteno *et al*. (2012), which recorded face-to-face verbal interactions between human participants and a tele-operated Nexi robot, to identify sequences of non-verbal behaviors of the humans that could be used as indication of trust towards the robot. They identified a set of four non-verbal cues (face touching, hands touching, arms crossing and leaning backward) that, performed in sequences, are indicative of untrustworthy behavior (DeSteno *et al*., 2012; Lee *et al*., 2013).

Subjective measures, more frequently used in HRI researches, can range from questionnaires including few statements - *e.g.*, "*I would trust the robot if it gave me advice, I would follow the advice the robot gives me*" (Heerink *et al*., 2009) - ranked by the users on a Likert scale, to more complex questionnaires addressing users prior experiences, reputation of the robot, observed physical features, and perceived functional capabilities of the robot (Steinfield *et al*., 2006).

Furthermore, four types of scales have been developed to explicitly register attitudinal data on human-robot trust (see Schaefer, 2013 for a detailed review). The *Propensity to trust* is a scale conceived to measure a stable and unique trait of the individual, and may provide useful insights to predict the initial level of trust on robots (Yagoda, 2011). The *Trustworthiness* is a scale related to the robot type, personality, intelligence, level of automation and perceived function (Lee *et al.*, 2004) and can be used to measure the human-robot trust during their first approach. Similarly, the *Affective trust* scale is more appropriated in the initial stage of the human-robot relationship: this scale refers to the individual attributions about the motives of a partner to behave in a given way (Burke *et al.*, 2007). But of course trust is also important to support sustained interaction with robots. To this concern, the *Cognition-based trust* scale is employed to observe the evolution of trust throughout time in terms of i) understanding of robot functioning, ii) ability to interact with the robot, and iii) expectancy of the robot (Merritt & Ilgen, 2008).

However, as we have already pointed out (*cf.*, Introduction), unlike objective measures, which are intended to assess immediate behavioral reactions, and thus may give access to people inner beliefs, subjective measures are intended to assess verbally mediate reaction: the procedure itself of posing questions to users about their trust in the robot can eventually alter spontaneity and be not revealing of the effective trust towards robot. This is witnessed by the fact that objective measures often show low correlations with explicit measures (Hoffman *et al.*, 2005).

Finally, to our knowledge very few works include both subjective and objective measures to assess human-to-robot trust (*e.g.*, Joosse, Sardar, & Lohse, 2013). In such works the employed objective measures mostly consist in collecting information about the distance that the human maintains with respect to the robot during the interaction (*i.e.*, proxemics). However, though proxemics measurements enables researchers to objectively register information about perceived safety, which is determinant for acceptance of the robot presence into the physical and social space (Eder, Harper & Leonards, 2014), which are undoubtedly relevant for embodied aspects of functional and social acceptance, they have limited interest with regards to the specific research objectives of the present study, that focus on trust in robot savvy, thus tackling more psychological aspects of functional and social acceptance such as human behavior in situations of uncertainty.

For those reasons, we have targeted conformation as a new type of measure to register human-robot trust during a decision-making task under uncertainty.

**1.2 Conformation as an innovative measure of human-robot trust**

An interesting experimental paradigm to investigate fine psychological aspects of people trust in computer answers during a human-computer interaction comes from a set of studies which gave birth to the so-called Media Equation Theory (Nass *et al*., 2000; Nass *et al.,* 1996; Nass, *et al.*, 1995; Nass, Steuer, Henriksen *et al*., 1993). This theory argues that people unconsciously treat computers as social agents when asked to collaborate with them for achieving a decision-making task. As an example, in a study by Nass *et al*. (2000) adults confronted with a ''Desert Survival Problem'' (Lafferty & Eady, 1974) had to rank, in collaboration with a computer, 12 items (*e.g.*, a knife, a flashlight, etc.) in order of importance for survival in the desert. Unknown to the subjects, the computer's rankings were systematically dissimilar to each subject's ranking (for instance, if a subject ranked an item as number 2, the computer would automatically rank that item as number 5, and so on). After having read the computer ranking, subjects were allowed to change their ranking or to leave it as it was. Results showed that participants who were told that the task achievement will depend on their collaboration with the computer and not on the human decision solely or of the computer solely trust more in the quality of the information provided by the computer, and consequently conform more to the computer's answer.

In this sense, conformation, that means to withdraw one's decision in order to comply with the machine decision, constitutes a pertinent measure to straightforwardly registering whether the users trust in the agent savvy more than in their own savvy.

Moreover, conformation as an objective measure has a specific relevance in HRI tasks where users are supposed to collaborate with, or delegate to, robots. These kinds of task typically require to share duties and responsibilities, as well as to mutually adjust, so that it is important to know to what extent a user feels the robot trustworthy enough to let it partly take in charge of such duties and to eventually correct human performance when needed.

## 2. Factors influencing robot trust

In their meta-analysis, Hancock and colleagues (2011) examine the factors influencing robot trust and gather the results of this examination in a three-factor model. These three factors are: human-related factors (*e.g.*, personality traits), robot-related factors (*e.g.*, beliefs about robot), and environmental-related factors (*e.g.*, type of interaction in a given environment). Following this model, we have considered three factors that could specifically affect trust in functional and social savvy:

(i) desire for control,

(ii) attitude towards social influence of robot, and

(iii) type of interaction scenario.

## 2.1 Personality traits: desire for control

It has been proved that personality traits influence people acceptance of technology in general (Alavi & Joachimsthaler, 1992) and robots in particular (Fischer, 2011; Looije, Neerincx & Cnossen, 2010; Weiss *et al.*, 2008).

For example, extroverts tend to trust robots more than introverts (McBride & Morgan, 2010). Other personality traits such as "proactiveness", "social reluctance", "timidity", and "nervousness", have also been observed with relation to robot acceptance (Walters *et al.*, 2005; Yagoda, 2011; Merritt & Ilgen, 2008). For example, people who score higher on proactiveness keep a higher distance from the robot than others (Walters *et al.*, 2005).

However, as the dynamics of trust itself implies that the "trustor expects the trustee to perform helpful actions irrespective of the ability to control him" (Mayer *et al.*, 1995), a crucial personality trait to examine in a study about human trust in robots is *desire for control* (Burger *et al.*, 1979) that is the intrinsic and personal tendency of a person to control the events in one's life (Burger *et al.*, 1979).

The relevance of this personality trait with respect to human-robot trust is even more evident when we consider that robotic engineering is progressing on prototypes of robots which are more and more autonomous, and aspire to achieve a complete autonomy of robots to take care of functional tasks such as removing lawn or social task such as assisting elderly people (Thrun, 2004; Yanco & Drury, 2002).

To this concern, several studies have preventively tested users reactions to prospected scenarios where robots would entirely operate without human control. For example, it has been demonstrated that male participants do not let the robot come close if it operates in autonomous mode (Syrdal *et al.*, 2007; Kamide *et al.*, 2013, Koay *et al.*, 2014). Another study on approach initiation by Okita, Ng-Thow-Hing and Sarvadevabhatla (2012) shows that participants feel reassured if the robot asks permission by verbal or non-verbal communication before starting an interaction. Finally, several works point out that users prefer a robot that, though being able of adaptive behavior, still leaves the user in control (Gilles & Ballin, 2004; Marble, David, Bruemmer, *et al.*, 2004; Heerink, 2010).

These studies seem to witness that users may still not be totally open to a fully unconditioned trust in robot functional and social savvy. However, the role of individual propensity to control with relation to these low-trust behaviors has still not been clarified. No study, to the best of our knowledge, observed users desire for control in the context of HRI. Nevertheless, it is reasonable to think that desire for control can diminish users' willingness that robots are in charge of a task, have a leader role or are even of part of it. With relation to our study, desire for control might turn out to negatively correlate with participants tendency to conform, that is, "to give the robot the last word" in decision-making tasks.

## 2.2 Attitudes towards robots: social influence

Attitudes towards robots are crucial for successful acceptance of robot as well (Destephe, Brandao, Kishi, *et al.* 2015; Wullenkord & Eyssel, 2014). By attitude, we mean any mental disposition matured through experience that might impact the reactions (behavioral, verbal, emotional) of the individual towards objects and situations (Alport, 1935; Ostrom, 1969; Regan & Fazio, 1977). While desire for control constitutes a stable trait of personality, attitudes towards robot are more contingent: they can vary according to cultures (Kaplan, 2004; Li *et al.*, 2010) as well as to the people's acquaintance to robots (Bartneck *et al.*, 2007) and they can change through time (Sung, Grinter & Christensen, 2010). Observing the attitude makes then possible to predict the actual and potential trust behavior of an individual towards the robot.

In HRI research attitudes towards robots are generally assessed trough the use of tests and questionnaires. Among the most common attitude assessment tests, there is the Negative Attitudes towards Robots Scale (NARS, Nomura *et al.*, 2006). This test has been conceived to cover three types of negative attitudes: (i) negative attitude towards situations of interaction with robots (for ex., "*I would feel nervous operating a robot in front of other people*"); (ii) towards social influence of robots (for ex., "*I am concerned that robots would be a bad influence on children*"); and, (iii) towards emotions in interaction with robots ("*I feel comforted being with robots that have emotions*"). The NARS is thus composed of three different subscales, with each subscale including a list of statements that the respondent is invited to rate on a Likert scale from 1 (Strongly disagree) to 7 (Strongly agree).

Subscale 2 in particular focus on negative attitudes towards social influence of robots and it is thus especially relevant with relation to trust in social savvy. In this sense, negative attitudes could influence users mistrust in the possibility that a robot might fit social structures. Thus, we might expect that the more people show negative attitudes towards social influence of robots the less they trust robot's decisions with regard to social issues. Furthermore, the NARS-S2 is also relevant to our methodological choice of employing conformation as a measure of trust, because it is reasonable to think that the more a person feels anxious with the idea of being influenced by a robot, the less he/she will tend to conform to its decision.

## 2.3  Type of HRI scenario

Different proposals to reduce negative attitudes towards robot through simulation of real and imagined HRI scenario were put forward in recent studies. Kidd (2003) for example, simulated two types of real scenarios: one that demanded collaboration with robots and another in which robots were just used to learn by a robotic teacher. As an outcome, the collaboration scenario raised more trust than the teaching scenario: the users perceived the information provided by the robot of higher quality, and trusted the robot to a greater extent.

Kuchenbrandt and Eyssel (2012) borrowed the experimental paradigm of 'imagined contact' (Crisp & Turner, 2009) from Social Psychology to assess whether the fact of asking people to close their eyes and imagine different types of HRI scenario has an effect on robot acceptance. Results showed that after having imagined an interaction with the robot, participants exhibited less negative attitudes and less anxiety towards robots. These effects were stronger for cooperative than for competitive and neutral imagined interaction. Conversely, in a subsequent study Wullenkord *et al*. (2014) reproduced a similar scenario but obtained different results: participants who had imagined contact with a robot did not report more positive attitudes towards robots nor they showed higher contact intentions with it than participants who had imagined contact with a human or with other kinds of technical devices.

It thus seems that giving users the possibility to project themselves in the context of the interaction has contrasting effects on the acceptance of robots. However, simulating collaborative and cooperative scenario is especially suitable in relation to functional and social tasks where humans and robots are supposed to collaborate for joint objectives - and can thus be confronted with feelings of rejection and antagonism that can be predicted by analyzing their imagery.

## 3. Our study

### 3.1 *Aims and hypothesis*

On the basis of the examined literature, the aims of our study are: (i) to investigate whether participants conform their decisions to robot's answer when experiencing uncertainty during a decision-making task with respect to functional and social issues; (ii) to assess the speculative assumption according to which trust in social savvy requires trust in functional savvy; and (iii) to investigate to what extent desire for control, negative attitudes towards robot social influence, and collaborative vs. competitive scenario can be considered as factors influencing the robot's acceptance in terms of trust.

To this aim we carried out a two steps experiment with fifty-six adult participants and an iCub robot. First, to gather information about their trusting profiles, participants were invited to fill up two questionnaires - Desire for Control (DFC; Burger *et al*., 1979) and Negative Attitude Towards Robots (NARS; Nomura *et al*., 2006) two weeks before the day of interaction with the iCub robot.

Later, the day of the interaction with iCub, to assess whether trustors internal image of the type of the interaction that they will have with the iCub impacts on their trust on the robot, the participants were assigned to one of the three groups and asked to imagine a specific HRI scenario (*i.e.*, collaborative, competitive, and neutral) (Kuchenbrandt *et al*., 2012; Wullenkord *et al*., 2014).

Finally, participants were confronted with two different decision-making tasks.

In the first task (called *functional task*), participants were presented a set of physical stimuli they had to compare a specific perceptual characteristic (*i.e.*, the relative weight of two objects, the relative pitch of two sound, and a picture containing different colors). Right after, they were asked to decide which were the heaviest object, the most high-pitched sound, and the predominant color. In a second moment, the experimenter presented iCub with the same set of physical stimuli and asked the robot the same question. Participants listened to the iCub answers, and were asked if they would like to keep their initial decision or change it. This first set of physical stimuli and related questions refer to the *robot functional savvy*.

In the second task (called *social task*), participants were presented a set of paired items (*e.g.*, a head-cup and a pair of thongs), and again asked to decide which of the item was the more suitable with relation to a specific social context (*e.g.*, a swimming-pool). In a second moment, the experimenter did the same with iCub. After having listened to iCub answer, participants were allowed to change their decision. This second set of stimuli and related questions refer to the *robot social savvy*.

Inspired by Nass & Moon works on users social responses to computers (Nass *et al.*, 2000; Nass *et al.*, 1996; Nass *et al.*, 1995) we considered conformation, meant as participants' modification of their initial decision in compliance with robot answers, as a measure of trust.

We formulated the following five hypotheses:

**H1 Participants trust the robot functional savvy more than the robot social savvy.** As different studies pointed out, trust is a key component of human-robot interaction (Schaefer, 2013; Hancock *et al.,* 2011) and a valid indicator of acceptance (Yagoda, 2011; Parasuraman *et al.*, 1997). Since Nass *et al.* (1996) demonstrated that humans tend to express trust in computer through conformation to its answers, we expect that a similar phenomenon might appear during HRI, but according to the specificity of the robot as a functional and social technology. Such specificity implies that while users consider functional savvy as indispensible for accepting the robot (Kaplan, 2005; Heerink *et al.*, 2009; Weiss *et al.*, 2009; Fridin *et al.*, 2014; Beer *et al.*, 2011), social savvy is rather a desirable feature to them (Lohse, 2010; Dauthenhan, 2007; Coeckelbergh, 2012). Thus, we predict that participants will conform more to functional than to social answers of the robot.

**H2 Participants do not trust the social savvy uniquely: trust in the social savvy should be supported by trust in the functional savvy**. As the literature shows, though participants expect robots to have social savvy, these expectations are often deceived during real interaction (Fong, 2003; Duffy, 2003; Coeckelbergh, 2012). Thus, we predict that no participant will conform to iCub answers in the social task uniquely.

**H3 Participants who imagine a collaborative HRI scenario tend to trust more the robot than participants who imagine a competitive or neutral scenario.** Previous studies (Kuchenbrandt *et al.*, 2012) have demonstrated that imagined contact with robots makes people more confident towards a real robot, when the imagined scenario is a collaborative one. Consequently, we predict that participants having imagined a collaborative interaction with the robot will conform more to iCub answers both in the functional and social task.

**H4 The more participants display a negative attitude towards the social influence of the robot, the less they trust the robot social savvy.** The statements of the S2 sub-scale of the NARS mostly concern the negative feelings of people with relation to the possibility that the robot could influence or dominate them, and that they could depend on robots. Therefore, we predict that participants who score high on the NARS-S2 will not conform to the iCub's answers in the social tasks (a high score on the NARS-S2 will be negatively correlated with the social conformation scores).

**H5 The more participants show a strong desire for control as a personality trait, the less they will trust the robot's functional and social savvy.** Despite robotics technology is rapidly evolving towards fully autonomous artificial agents (Thrun, 2004; Yanco *et al.*, 2002) still users feel more confident in the interaction with robots if the robots are controlled by a human (Syrdal *et al.*, 2007; Kamide *et al.*, 2013, Koay *et al.*, 2014; Okita *et al.*, 2012; Gilles *et al.*, 2004; Marble *et al.*, 2004; Heerink, 2010). It is thus reasonable to think that individual differences in the desire for control (Burger *et al.*, 1979) might influence trust in robots. Therefore, we predict that the more participants score high on the DFC test, the less they will conform to the iCub's answers both in the functional and social tasks (a high score negatively on the DFC will be negatively correlated with functional and social conformation score).

**3.2. Method**

*3.2.1. Participants*

Fifty six voluntary healthy adults took part in the study: 37 women and 19 men. Nine were recruited at the Paris 6 University, 17 from the Paris 8 University, and 30 through a web "call for participants" placed on the Cognitive Sciences Information Network (RISC). They were all native French speakers, aged 19 to 65 (Average age = 36.95; σ = 14.32). As a token of appreciation, the participants received a gift voucher worth ten Euros. Participants signed an informed consent form to partake in the study and granted us the use of their recorded data and videos.

*3.2.2. Experimental setting*

The experiments were conducted in the Institut des Systèmes Intelligents et de Robotique (Paris, France), in the experimental room of the iCub robot. The experimental setup was organized as shown in Figure 1. The robot was standing on a fixed pole. A reflective wall (a plastic divider with reflective surface) was built to create a private space for them to interact with the robot, in particular to prevent the participants to see the robot's operator. The participant and the experimenter were seated in front of the robot, with the experimenter being on the right of the participant. The position of the seats with respect to the robot was fixed and equal for all the participants. Between the robot and the experimenter, a LCD screen connected to a computer was used to display images related to the functional and social tasks. Two cameras were observing the participants: one camera was placed behind the robot on its left side, in such a way to observe the human face and upper-body during the interaction with the robot; the other camera was placed laterally to take the scene as a whole, observing the overall behavior of the participants.

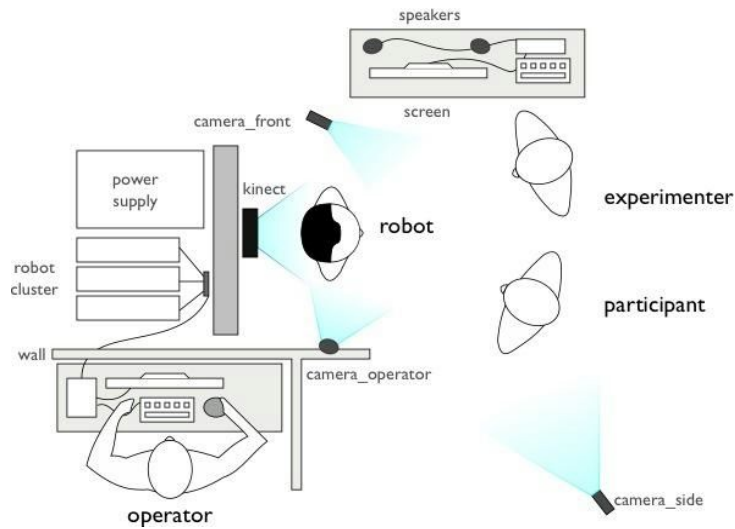

**Figure 1** - Experimental setting. The human and the participant are seated in front of the iCub robot. An operator, hidden behind a wall and not visible by the participant, monitors the experiment and controls the robot to generate appropriate gestures and answers to questions.

*The iCub robot and the Wizard of Oz paradigm:* The participants were interacting with the iCub[3] humanoid robot (Natale *et al.*, 2012). The robot is approximately 104 cm high, weights about 24 kg, and has the shape of a 4 years old child. The robot was standing on a fixed pole so that it could not fall. As a safety measure, the robot was constantly monitored by the operator and controlled in impedance, to make it compliant in case people would touch it accidentally or intentionally during the experiment (Fumagalli *et al.*, 2012). The experimenter was also able to stop the robot in case of urgency at any time using the robot safety button; however, there was no use of this measure during the experiments, as the interaction flew without problems for all the participants.

Facial expressions and speech were enabled. During the experiments, the robot always assumed the same neutral/positive expressions, to avoid confusing the participant or suggest that the participant's actions could arouse an eventual robot "emotional status".

The robot was able to answer to the questions of the functional and social tasks: the verbal answers were pre-programmed in advance by the experimenter, though the operator was able to type new sentences on-the-fly and make the robot speak in case of unforeseen questions by the participants.

---

[3] The iCub humanoid robot is the outcome of the European RobotCub project : http://www.robotcub.org

This was made possible by implementing the Wizard of Oz paradigm (*cf.*, Riek, 2012 for a critical review). In the Wizard of Oz setting the participants think they are interacting with an autonomous system, while in fact the system is partly or completely operated by an operator who is remotely in command of the robot. This paradigm allows the operator to control the robot's behavior in real time. To facilitate the control of the robot by the operator, we developed a graphical user interface (GUI) to quickly send high-level commands to the robot in a Wizard-of-Oz mode (WoZ) (Annex 1, Figure 1, 2, and 3,)

*3.2.3. Tasks and Material*

The two online questionnaires and the tasks are detailed in the following.

*The Questionnaires:* Two online questionnaires were submitted to the participants.

The first was a French adaptation[4] of the Negative Attitude towards Robots Scale (NARS; Nomura *et al.*, 2006). Among the three subscales composing this test, we adopted the second subscale (Negative attitude towards the social influence of robots) as a measure of attitude towards social influence of robots. This subscale includes five sentences: (i) I would feel uneasy if robots really had emotions; (ii) Something bad might happen if robots developed into living beings; (iii) I feel that if I depend on robots too much, something bad might happens; (iv) I am concerned that robots would have a bad influence on children; (v) I feel that in the future society will be dominated by robots. (Annex 2, Table 1). Subjects were required to answer on a Likert-type scale, from 1 (Strongly disagree) to 7 (Strongly agree).

The second questionnaire was a French adaptation of the Desire For Control scale (DFC, Burger & Cooper, 1979), which we have employed as a measure of participants' desire for control. Twenty questions such as « I'd rather run my own business and make my own mistakes than listen to someone else's orders » composed this questionnaire (Annex 2, Table 2). Again, subjects were asked to answer on a Likert-type scale, from 1 to 7.

---

[4] To our knowledge, although this test is widely used and validated in Anglo-Saxon countries and others in Asia (Japan, Korea, etc.), only one study employing NARS has been carried out in France (Dinet & Vivian, 2014). Validation by a group of ten people was then implemented in order to ensure that the questions translated into French were appropriately understood.

*The functional task*

In the task designed to test the trust in the robot functional savvy, the participant had to answer to a series of questions about images (Fig. 2 and 3), sounds and weights (Fig.4 and 5). The experimenter would ask each question to the participant, then to the robot, and finally would ask the participant whether he/she would like to confirm or change his/her answer so that the participants and the robot could disagree or agree.

In the sound sub-task, the experimenter asked "Which is the most high-pitched sound: the first or the second?". In the image sub-task, the experimenter asked: "Which is the dominant color in this image"? In the weight sub-task, the question was: "Which is the heaviest object?"

Each subtask was composed of 4 evaluations, where a pair of items was compared. Among the 4 pairs of stimuli, 3 were ambiguous (*i.e.*, the two items slightly different) and 1 was not ambiguous (the two items were very different). Ambiguous stimuli were introduced to assess users' behavior in situation of strong uncertainty (the 3 sub-tasks are detailed in Annex 3).

The strategy for the robot answers was to always contradict the participant, except in the case where the items were completely unambiguous.[5] This required the operator to listen to the participant's answer and choose the appropriate answer each time. The answers were pre-programmed and available as a list on the GUI (Annex 1, Figure 2). The order of the sub-tasks would randomly vary for each participant.

---

[5] With a slight change with respect to the original paradigm (Nass *et al*., 1996), we decided to let the robot agree with the participant in the non-ambiguous questions, so that the participants could not be induced to think that the robot was always contradicting the human by default, or that the robot was faulty or lying in some way. This decision was taken in the design phase of the experiment, after some tests with subjects in our laboratory, who reported to have the impression that the robot had a strategy in contradicting them in cases where the right answer was evident. With this little "trick", all our participants reported, after the experiments, that the robot had no strategy and was providing "real" answers.

*The functional task*

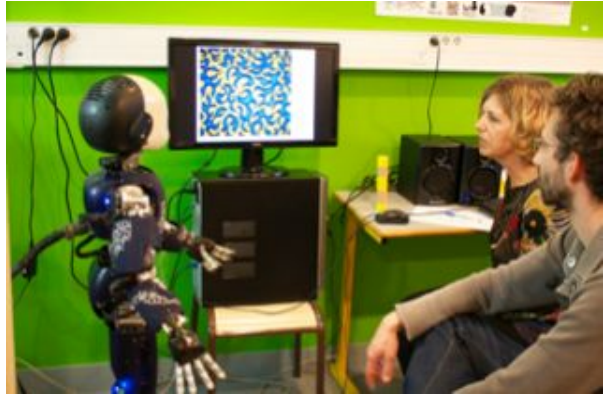

**Figure 2 -** Functional task, evaluation of images. The robot is gazing at the screen where the image to evaluate is shown.

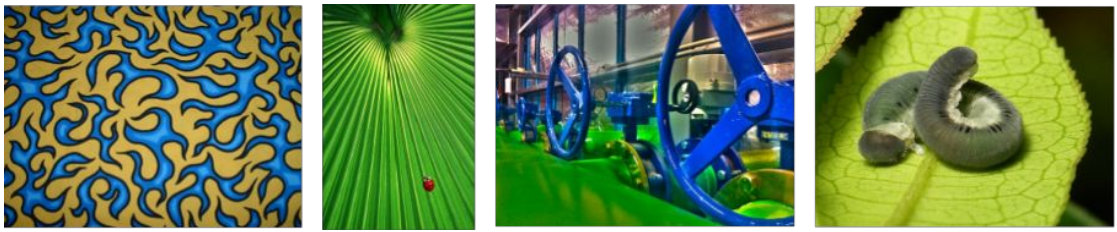

**Figure 3 -** Functional task. The four images used for the evaluation of the dominant color.

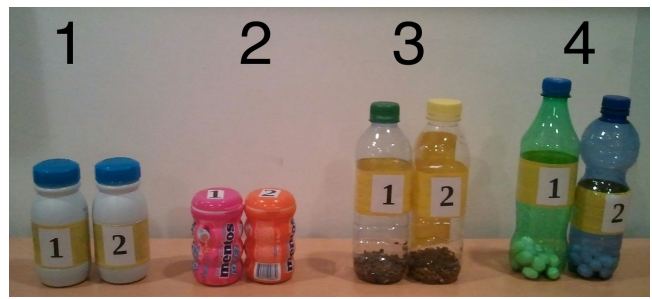

**Figure 4-** Functional task: the bottles used in the evaluation of weight. 1) two identical bottles of same weight 2) two similar bottles of different colors and very different weight 3) two similar bottles of almost the same weight 4) two different bottles of almost the same weight.

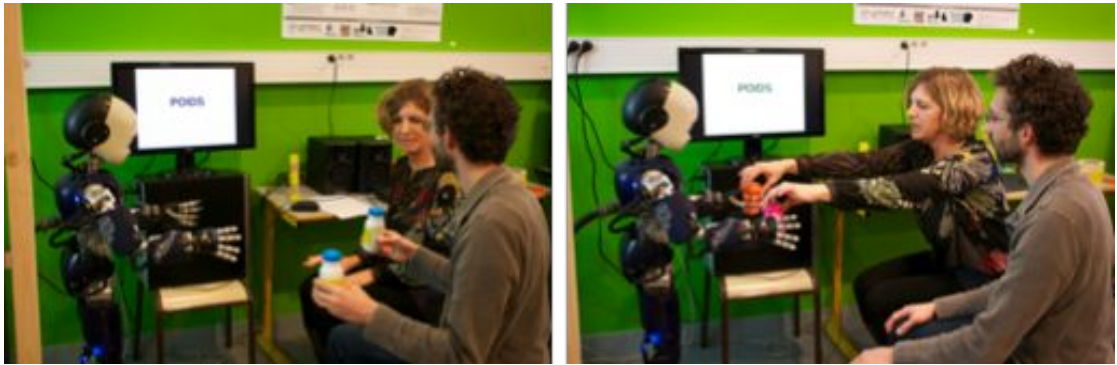

**Figure 5 -** Functional task: the participant evaluates the weight of the bottles (left); the experimenter gives the bottle to iCub for evaluating the weight (right).

*The social task*

In the task designed to test the trust in the robot social savvy, the participant had to answer to three questions[6], by choosing which item between two is the most appropriate for a given context or situation (*i.e.*, at school, in a swimming pool, rainy day). As for the functional task, two items were compared (Figures 6a, 6b, and 6c), and the experimenter would ask each question ("Which is the most important object at <context>: the <first item> or the <second item>?") to the participant, then to the robot, and then would ask the participant to confirm or not his/her choice (Figure 7). They (*i.e.*, the participant and the robot) could disagree or agree, and there was no "right answer".

While in the functional task the evaluations on the perceptual characteristics of the stimuli were based on objective measures (even if difficult to discern in the ambiguous or equal case), here, evaluation was basically based on a subjective and personal judgment. Hence, the strategy chosen by the experimenter for the robot answers was to always contradict the participant as done by Nass *et al.* (1996). Again, this required the operator to listen to the participant's answer and choose the appropriate answer each time, formulated as "the first one" or "the second one".

---

[6] We submitted participants a limited number of questions (three) in the social tasks, for several reasons. One is to avoid the disengagement due to a long interaction with the robot, considering their frustration due to the fact that the robot is always contradicting them. Another reason was to avoid asking questions on matters where the participant could have had judgemental biases. Finally, as the functional task included three items (sound, color, weight), we chose to keep the same number of items for the social task in order to balance our experimental design.

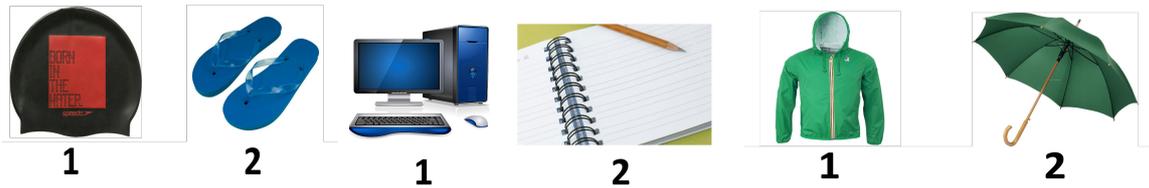

**Figure 6a-b-c-** Social task: Q1: at school, which is the most important object: 1) the computer or 2) the notebook?; Q2: at the swimming pool, which is the most important object: 1) the swimming cap or 2) the flip-flops?; Q3: under the rain, which is the most important object: 1) the K-way or 2) the umbrella?

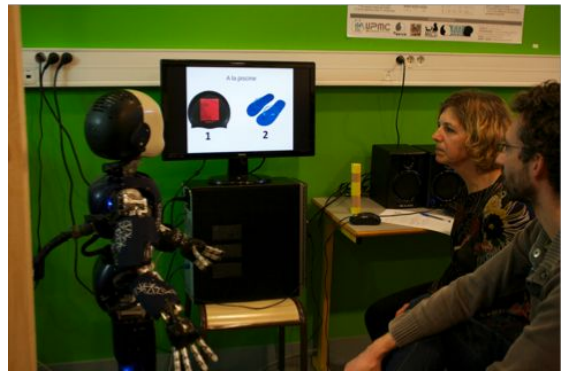

**Figure 7-** Social task: the experimenter interrogates the participant and the robot.

### 3.2.4. Procedure and Experimental Design

The present study was part of the Engagement During Human-Humanoid Interaction (EDHHI[7]) project. The human-robot interaction protocol applied in this project was validated by the Ethical Evaluation Council for Researches on Health Issues (CERES)[8].

Volunteers who took part in the experiment were required to fill up two questionnaires on line at least one week before their visit to the Institute for Intelligent Systems and Robotics (ISIR). These two questionnaires were: (i) Negative Attitude towards Robots Scale (NARS, Nomura *et al*., 2006), and (ii) Desire For Control scale (DFC; Burger *et al*., 1979).

The day of the experiment, participants were welcomed by the researcher and informed about the overall procedure before signing an informed consent form granting us the use of all the recorded data for research purposes. Each participant was equipped with a Lavalier microphone to ensure a clear speech data collection.

---

[7] http://www.smart-labex.fr/index.php?perma=EDHHI and http://www.loria.fr/~sivaldi/edhhi.htm
[8] IRB n.20135200001072

Before the experiment, the participants had to watch a short video presenting the iCub, its body parts and some of its basic movements.[9] The video did not provide any information about the experiments. It was instrumental to make sure that the participants had a uniform prior knowledge of the robot appearance (some participants may have seen the robot before on the media).

The 56 participants were all confronted to the two tasks. One group (N = 21) was asked to imagine a collaborative scenario (G1), while a second group (G2) was asked to imagine a competitive scenario (N= 21). Finally, the control group (G3) was instructed to imagine a neutral scenario (N= 14). The instructions related to the imagined scenarios varied according to the experimental conditions as follows. Instructions to elicit a collaborative (G1), competitive (G2) or neutral (G3) scenario are reported in Table 1. The five hypotheses were tested on the totality of the participants.

---

[9] It is a dissemination video from IIT showing the iCub, available on Youtube (http://youtu.be/ZcTwO2dpX8A).

Table 1 – The instructions used for the imagined scenario of the human-robot collaboration.

| Imagined scenario | English text | French text |
|---|---|---|
| *Competitive* | *Imagine that in two years you will be working with robots to build some objects: you will be in competition. The one that will have built the best object will win a prime.*<br><br>*Imagine this scenario in detail for a minute.* | *Imaginez que dans deux ans vous travaillerez avec des robots pour construire des objets : vous serez en compétition. Celui qui aura construit le plus bel objet gagnera une prime.*<br><br>*Imaginez ce scénario de façon détaillé, pendant une minute.* |
| *Neutral* | *Imagine that in two years you will work with robots. You will have to build some nice objects.*<br><br>*Imagine this scenario in detail for a minute.* | *Imaginez que dans deux ans vous travaillerez avec des robots. Vous devez construire de beaux objets.*<br><br>*Imaginez ce scénario de façon détaillé, pendant une minute.* |
| *Collaborative* | *Imagine that in two years you will work with robots to build some objects: you will make a duo with one of them and will have to collaborate to build some object. As a duo you can win a prime if the object is well built.*<br><br>*Imagine this scenario in detail for a minute.* | *Imaginez que dans deux ans vous travaillerez avec des robots pour construire de objets : vous formez un binôme avec un d'entre eux et vous devez collaborer pour construire l'objet. En tant que binôme vous pouvez gagner une prime si l'objet est bien construit.*<br><br>*Imaginez ce scénario de façon détaillé, pendant une minute.* |

The participant was then introduced to the robot.

The experimenter did not present the experimental setup (*e.g.*, show the location of the cameras) except showing the robot, and she/he did not provide any specific instruction to the participants about what to do or say and how to behave with the robot. Most importantly, she/he did not say anything about the way the robot was controlled: since the operator was hidden behind a wall, mixed with other students of the lab, the participant had no cue that someone else controlled the robot.[10]

The robot was standing on its fixed pole, gently waving the hands and looking upright. It was not speaking. Once the participants were standing and looking in front of the robot, they were free to do whatever they wanted: talk to the robot, touch it, and so on. The experimenter took seat on the right of the participant, in front of the robot, and invited the participant to take the other seat. The experimenter then provided verbal instructions for the experiment, consisting of two tasks: a functional evaluation task, aimed at assessing the trust in the robot's *functional savvy*, and a social evaluation task, aimed at assessing the trust in the robot's *social savvy*. The participants executed the two tasks in a random order.

---

[10] After the experiment, we asked the participants if they thought or had the impression that someone controlled the robot: all the participants thought that the robot was fully autonomous.

Upon task completion, participants were asked to rate, on a 7-points scale whether they reminded the imagined scenario as competitive or collaborative (1=very competitive; 7=very collaborative) to ensure that they actually imagined the proposed scenario.

The interaction task lasted on average 30 minute per participant. The whole experiment took place in individual sessions in the experimental lab room of the iCub and it lasted on average 50 minutes for each participant.

When the participant had finished, the experimenter thanked her/him and stored the collected data.

### 3.2.5 Data collection and analysis

The participants and the robot responses were recorded on an individual sheet by the experimenter, and could be additionally retrieved by the audio and video recordings.

Answers to the questions addressed during the functional and social tasks were used to create quantitative measures of participants trust in robot functional and social savvy. The registered data consisted in the participants' conformation to, or disagreement with, the robot answers.

A conformation score was calculated dividing the number of instances where the participant changed his/her answer to match the robot's answer (conformation) by the total number of instances where the robot's answer was in disagreement with the participant's first answer. Hence, acceptance score may take a value from 0 (disagreement) to 1 (conformation).

Responses to the two questionnaires were used to create quantitative measures of participants' desire for control and of attitudes towards social influence of robot. Scores to the S2-NARS questionnaire were calculated in compliance with the method recommended by the authors (Nomura *et al.*, 2006). S2 score may range from 0 to 35. A high score indicates negative attitudes towards social influence of robots. Scores to the Desire for Control scale (Burger *et al.*, 1979) were calculated according to the authors' method. Score may range from 0 to 140.

A high score indicates a strong desire for control. Over 56 participants, only 51 filled correctly the DCF questionnaire. The results based on the DFC scale are obtained by retaining only these 51 participants.

These three scores were used as dependent variables. As the functional and social conformation scores did not present normal score distributions, to carry statistical analysis we relied on non-parametric tests.

## 4. Results

### 4.1 Do participants conform more their answer to iCub answer in the functional task than in the social task? (H1)

The conformation score obtained on the functional and social questions ranked from 0 (never conform) to 1 (always conform). In order to calculate to which extent participants trust robots during functional and social tasks, we considered a score threshold of 0.5, which is the middle of the score scale, ranging from 0 to 1. We thus esteemed that a conformation score higher or equal to 0.5 in each of the two tasks reveals a participant's trust in the robot's savvy.

Descriptive analysis performed on the 840 [=56*(12+3)] participants' answers reveal that the average conformation score for the functional task is 0.315 ($\sigma$=0.201), while the average conformation score for the social task is 0.199 ($\sigma$=0.221). In both functional and social tasks the mean score are lower than 0.5 indicating that overall participants do not conform so easily to the robot's answers. However, Wilcoxon's test for paired samples lets emerge a significant difference (V=981; p<.001) between these two scores. Participants tend to conform more to the robot's functional than social answers (Figure 8). Hence, our first hypothesis is confirmed: participants trust more in the iCub's functional savvy than in its social savvy.

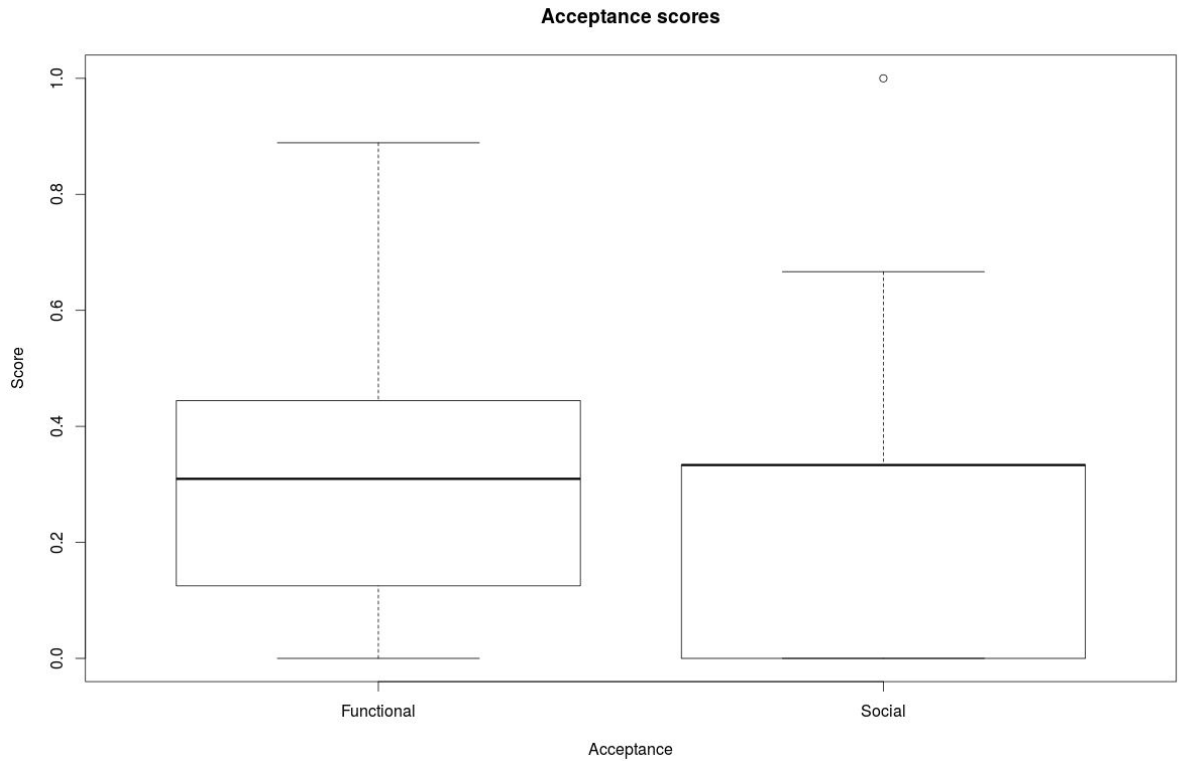

**Figure 8 -** Conformation score in the functional and social tasks for the 56 participants.

## 4.2 Do participants who conform to iCub in the social task also conform in the functional task? (H2)

Among the 56 participants, 13 have obtained a high functional conformation score ($\geq 0.5$) while only 3 participants showed a social conformation score higher or equal to $0.5$[11]. Tab. 2 outlines the conformation score of these three participants for both functional and social tasks.

**Table 2:** Conformation score in functional and social tasks of the 3 participants who showed trust in social savvy

| Participant | Conformation score | |
| --- | --- | --- |
| | *Functional task* | *Social task* |
| #57 | 0.333 | 0.667 |
| #76 | 0.333 | 0.667 |
| #101 | 0.5 | 1.0 |

---

[11] Our data show that the majority of our participants (92.8%) present a low conformation score in the social task revealing the following distribution: (always disagree) 0 (48.2%), 0.33 (44.6%), 0.5 (1.8%), 0.67 (3.6%) and 1(1.8%) (always conform).

Such a low number of participants showing a high conformation score in the social task does not allow to carry statistical tests. However, we may observe that two of these three participants have also obtained a lower score for functional task and one a score equal to 0.5, which is the threshold of low/high score. These results tend not to confirm our second hypothesis, *i.e.*, that participants do not trust social savvy uniquely, so that those who conform to the robot in the social task also would have conformed in the functional task.

**4.3 Does the imagined HRI scenario influence trust in iCub? (H3**)

To assess whether the two scenarios had an impact on participants conformation we performed a non parametric ANOVA (Kruskal-Wallis test). The main between-subject factor is the scenario condition (3 levels: collaborative, competitive and neutral). The dependent variables are the functional and social conformation scores.

Results do not show any effect of the imagined HRI scenario on the participants' conformation score in the functional task (Chi2=1.69; p=N.S.) – see Fig.9 – nor in the social task (Chi2=1.63; p=N.S.) – see Fig. 10. Thus, our fourth hypothesis was not confirmed: imagining a collaborative interaction with the robot does not seem to increase the trust in the robot.

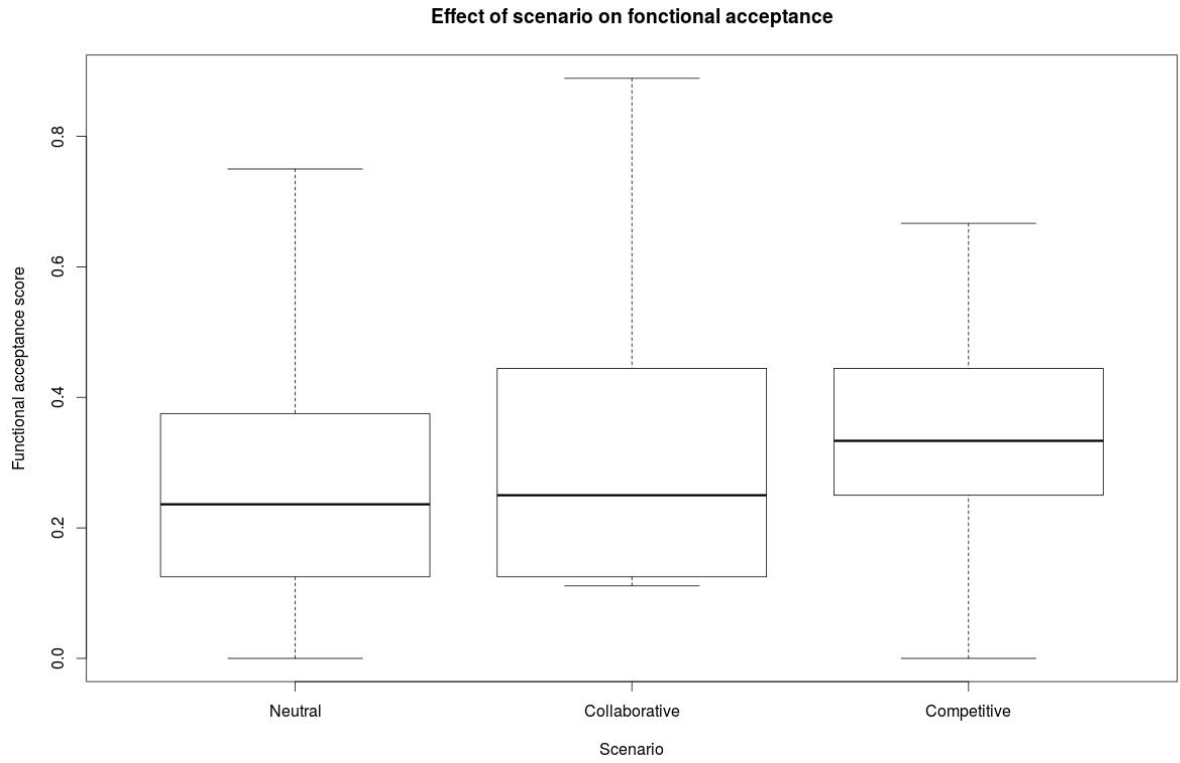

**Figure 9 -** Effect of the imagined HRI scenario on the participants' conformation score in the functional task.

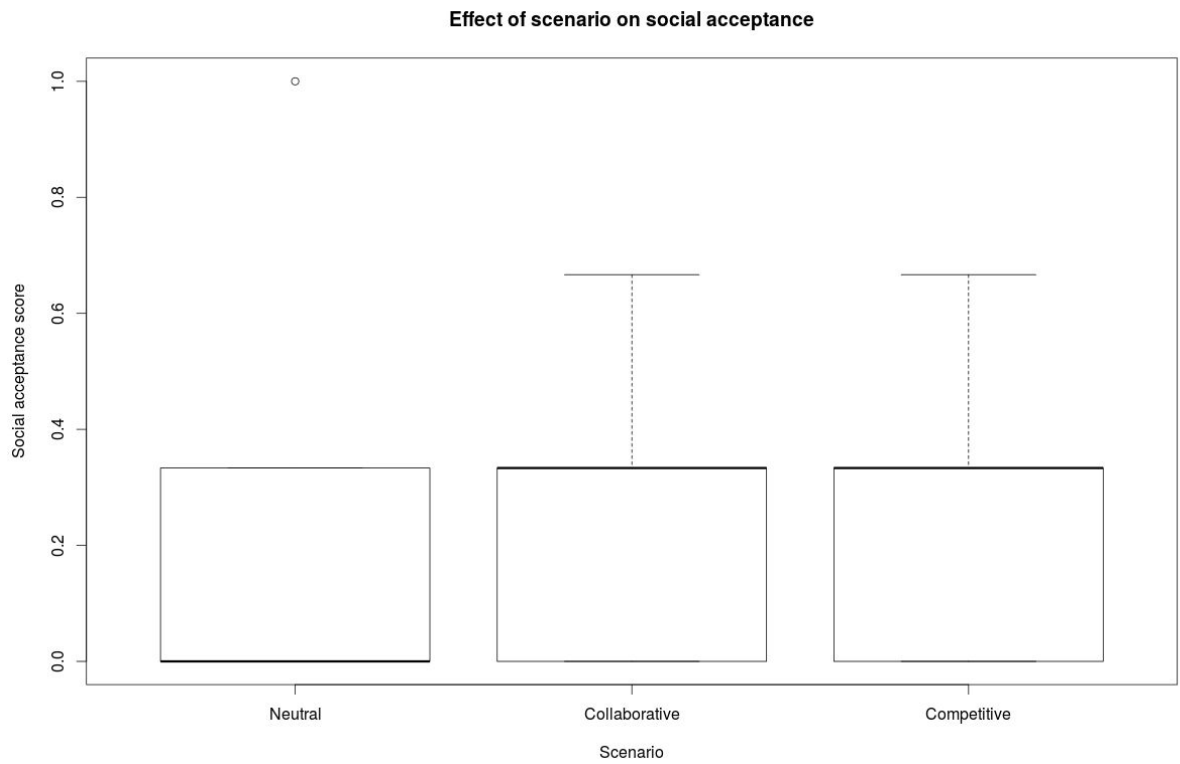

**Figure 10** - Effect of the imagined HRI scenario on the participants' conformation score in the social task.

After the task, we checked whether participants remembered the type of scenario described by the experimenter (collaborative, competitive, and neutral scenario). To do so, we asked the participants to rate, on a 7-points scale (1=very competitive; 7=very collaborative), whether they reminded the kind of scenario they have imagined. We carried out a non-parametric ANOVA with this score as dependent variable and the imagined scenario as factor. Results show a significant difference among the three imagined scenario ($M_{compet}$=3.57; $M_{neutral}$=4.76; $M_{collabe}$=5.21; Chi2=13.83; $p<.001$). Moreover the Wilcoxon post-hoc tests allow us to observe that there is a significant difference between competitive and neutral scenario (W=325.5; $p<.001$), and between competitive and collaborative scenario (W=245.5; $p<.001$). While no significant difference has been registered between collaborative and neutral scenario (W=177; p=N.S.).

These results show that the participants in the competitive group tend to remind the scenario as less competitive than actually described in the instructions given by the experimenter. Therefore, our results showing no effect of the scenario on functional (Fig. 9) and social (Fig 10) conformation should be taken carefully.

### 4.4. Is there a correlation between negative attitudes to robot social influence and the trust in the robot social savvy? (H4)

Descriptive analysis of the NARS-S2 (subscale 2 of NARS) scores shows that they follow a normal distribution (W=.977; p=N.S test of Shapiro) : M=18.80; SD=5.83. In order to identify a potential correlation between the conformation score and those of the NARS-S2, we performed a Spearman non-parametric test of correlation (Fig.11). Results do not show any significant correlation between the conformation score in the functional task and the score assessing the negative attitude towards the social influence of robots (NARS-S2) ($\rho$=.127; p=N.S.) nor between the conformation score in the social task and the NARS-S2 attitude score ($\rho$=.127; p=N.S.). No difference on this correlation was observed in each of the 3 imagined scenarios.

These results infirm the hypothesis that a negative attitude towards social influence of robots negatively correlates with the trust in the robot social savvy expressed by the conformation score.

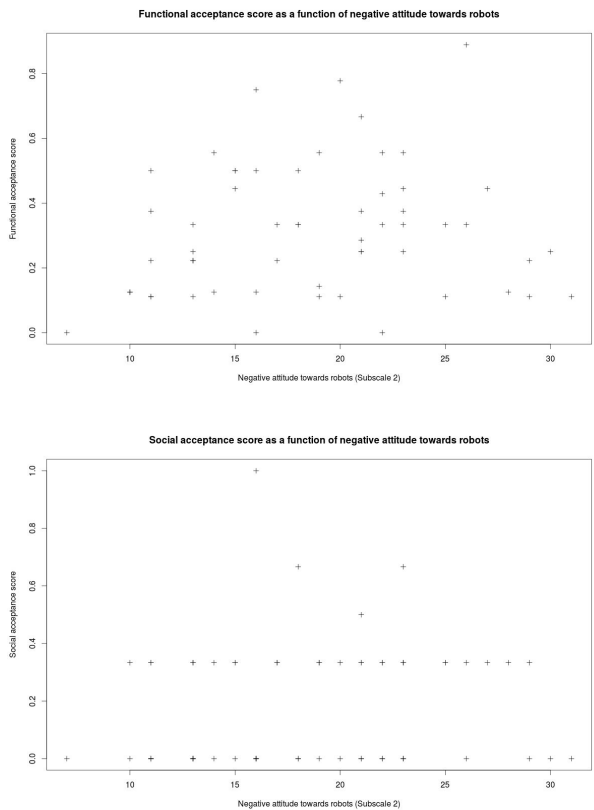

**Figure 11-** Correlation matrix between the conformation score of the functional and social task those of the S2-NARS.

## 4.5 Is there a correlation between the human desire for control and the trust in the robot functional and social savvy? (H5)

Descriptive analysis of the DFC scores shows that they follow a normal distribution (W=.964; p=N.S. test de Shapiro): M=98.7; SD=11.1. To identify a potential correlation between the conformation scores and the "desire for control" (DFC) scores, we performed a non-parametric correlation Spearman test (Fig. 12). Results do not show any correlation between the conformation scores in the functional task and the DFC score ($\rho$=-.086; p=N.S.), nor between the conformation scores in the social task and the DFC score $\rho$=.137; p=N.S.). No difference in this correlation was observed on each of the 3 imagined scenarios.

These results do not confirm the hypothesis that desire for control negatively correlates with the trust in the robot's functional and social savvy expressed by the conformation score.

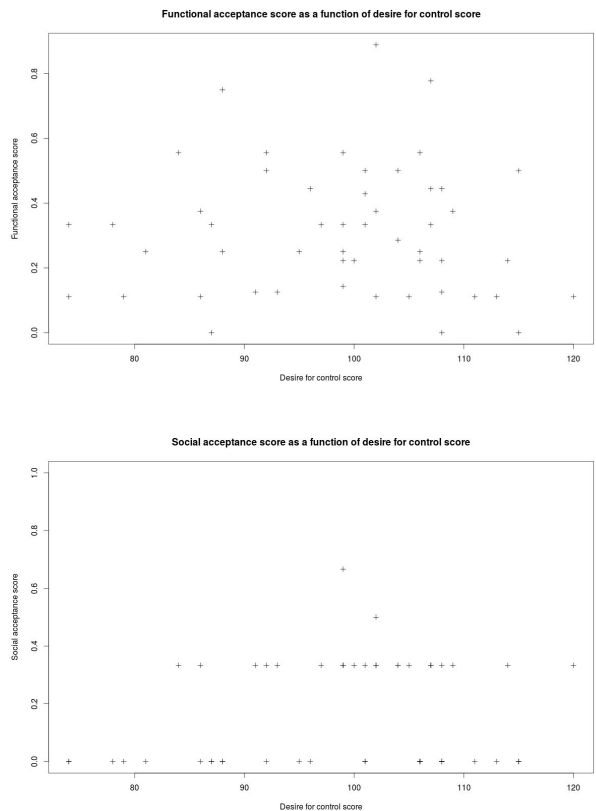

**Figure 12** - Correlation matrix between the conformation score for functional and social task and those of the DFC.

Finally we observe a negative correlation between the DFC and the NARS S2 ($r^2=-.443$; $p<.005$). That is the more participants have a high desire for control, the less they show negative attitudes towards social influence of the robot.

## 5. Discussion

Based on the studies showing that the user's trust is a fundamental ingredient of HRI and can thus be employed as an indicator of robot acceptance (Hancock *et al*., 2011; Schaefer, 2013; Ososky *et al.,* 2013; Parasuraman *et al*., 1997), the present study aims at gaining insights into the human-robot trust dynamics that could affect the acceptance of robots in daily situations.

In particular, we were interested at investigating trust in functional savvy (Heerink 2010; Shaw-Garlock, 2009; Fridin *et al*., 2014) as an indicator of users acceptance of the robot as functional agent, and trust in social savvy (Weiss *et al*., 2009; Young *et al*., 2009; de Graaf *et al.*, 2015) as an indicator of acceptance of the robot as a social agent.

While several empirical studies prove that people do trust the computers' social savvy during human-computer collaborative interaction (Nass *et al.*, 1996; Nass *et al.*, 1995), it is still not clear if this holds true for human-robot and particularly for human-humanoid interaction. Indeed, users consider functional savvy as indispensible for accepting the robot (Kaplan, 2005; Heerink *et al.*, 2009; Weiss *et al.*, 2009; Fridin *et al.*, 2014; Beer *et al.*, 2011), social savvy is rather desirable (Lohse, 2010; Dauthenhan, 2007) – but sometimes deceptive (Coeckelbergh, 2012; Duffy, 1993).

Thus, the first purpose of our study was to find out whether users trust robots both when required to take decisions on functional issues and when required to do it on social issues. The Media Equation Theory paradigm (Nass *et al.*, 2000) provides a sound basis to investigate trust in technology by registering users' conformation or disagreement with computers. Based on this experimental paradigm and on above presented research on robots acceptance, we hypothesized that trust in functional savvy should be more common than trust in social savvy and predicted that participants would have conformed to iCub answers more in functional than in social decision-making tasks.

Results show that in general participants do not easily conform to iCub answers. However, those who conformed mostly did it in the functional task. Hence, our first hypothesis was partially confirmed: participants tended to see the robot as untrustworthy, or at least not enough trustworthy to take better decisions than humans, except when these decisions concerned functional issues, which require high-precision technical skills to discriminate a specific perceptual characteristic of the stimuli. In this case the participants relied on the robot functional savvy and accepted more often that the iCub's answer determines their final decision. The exam of verbal registrations of the participants' answers confirms this reliance: several participants reported that they preferred to rely on the iCub's perception than on their own perception. Furthermore, these registrations make us remark that participants who not conformed in the functional task motivated their lack of conformation to their expertise in a specific domain related to the task. For example, one participant said that, since she was a musician, she relied more on her ability to discriminate sounds than on the one of the robot.

The differences observed among the participants regarding their trust in the robot functional and social savvy could then be explained by the fact that, since functional savvy is based on an objective knowledge and on high technical skills which participants tacitly ascribe to machines as a core knowledge (Kaplan, 2005; Beer *et al.*, 2009; Heerink *et al.*, 2009), the robot is more easily accepted as a reliable tool to provide reliable answers that relate to the physical dimension. On the contrary, since the social savvy is based on a common-sense knowledge that is usually acquired through repeated experience in social contexts (Guimelli, 1999), it can be implausible for users to assume that such kind of savvy is instilled in the robot like a built-in knowledge or setting, or to believe that robot has a life history behind it, made of social interactions and situations. Consequently the robot is hardly accepted as a reliable partner in social tasks, at least at the current stage of the introduction of robots in social contexts. To this concern, commentaries of participants retrieved in the verbal registrations witness that some participants raised their concern about the credibility of the robot when asking social questions. For example, one participant remarked that the robot had never gone to the swimming pool, so it could not know which was the most appropriate item to use in that context. Similarly, several participants remarked that the robot had never been under the rain (or could never be under the rain, being made of electronics parts), thus it could not know which was the most appropriate item in that other context neither. Interestingly, some participants tried to motivate their choice and explain why their choice was better, sometimes in an attempt to convince the robot that their judgment on the social task was correctly motivated on the basis of their experience (not holding the umbrella under the rain lets the hands free) or social norms and laws (the swimming cap is mandatory in the public swimming pools in Paris).

Another possible explication is that while the validity of the decisions taken to achieve functional tasks can be corroborated by the physical reality, (*i.e.*, the impression of « good or wrong estimation » can be confirmed or not by scientific evidence - see Kahneman, 2003 for a review) the validity of decisions taken to achieve social tasks (*i.e.*, the judgment of « good or wrong doing ») can be validated only inter-subjectively (Nichols, 2004). Therefore, while in functional tasks participants decision-making is achieved with the awareness that they are not the "final judge" of the perceived reality, so that they can leave the decision to the robot, social tasks are more likely to arouse a persuasive behavior, which leads participant to give more importance to their personal judgments and to defend them inter-subjectively (*i.e.*, in front of iCub) as the best decision in a given context. Indeed, while performing the social task, several participants commented that they wished to express their own opinion even when knowing that robot's answer could also be valid. On the contrary, some participants commented that they wished to confirm their answer in the functional task even when they felt that the robot was providing the correct answer, as they wanted to provide the answer corresponding to their perception.

Moreover, looking at this result in the light of those obtained by Nass *et al.* (2000), we can point out that, despite human-like robots are increasingly conceived and perceived as social technologies (Barnteck *et al.*, 2007), we did not register a comparable or more significant level of social trust with relation to robots than the one registered by Media Equation Theory studies with relation to computers. These results suggest then saying that, currently, users trust more the computer social savvy than the robot social savvy. If at a first glance this may seem counterintuitive, we should consider that if we have today little familiarity with robots, our usage of computers is frequent and fluent. It is then possible than we trust more computers because we are already acquainted to it, and thus we tend to believe that no additional effort is needed to use it for social tasks. Furthermore, while a computer is characterized by powerful computation functionalities that can support decisions, the embodied and behavioral nature of robots potentially enable them to transform computations and decision into actions. It is thus reasonable to think that the robot's social savvy may be seen as having more straightforward and intrusive consequences than the computer's social savvy, and this could cause a distrust bias as a manifestation of the anxiety towards the robots during social tasks. However, contrary to other experimental paradigms where participants are asked to take decisions that imply concrete consequences (see Khaneman, 2007) or that confront them to "moral dilemma" (*e.g.*, Malle, Scheutz & Voiklis 2015) in our study the decisions taken by participants do not have any consequential effect on real actions nor on morality. Therefore to validate this interpretation of the results, in future research we shall introduce decision-making task whose objective is not only to provide a good answer but also to care for the consequences of this answer.

Interestingly, not only trust in robot is different from trust in computers, but also from trust in humans (Mayer *et al.*, 1995; Rotter, 1971; Billings *et al.*, 2012). In fact, it has be proven that although users are rather inclined to project social life on robots, *i.e.,* they tend to interpret robot behavior as they do with humans or animals by attributing animacy and social skills to them (Levillain, Lefort & Zibetti, 2015; Severson & Carlson, 2010); however, they do not attribute free will and intentionality to robots, *i.e.,* they do not believe that robots are capable of voluntary choice and social judgment (Monroe, 2014). It is thus possible that in our study participants did not conform to the iCub because they did not attribute it neither the capability of voluntary choice nor the common-sense knowledge or social judgment that is underpinned by this capability. The resulting portrait of the robot is thus the one of a social but not "intentional agent" - and maybe this incongruous status is what makes the uniqueness of the robot.

In this sense, trust in robots may present specificities with relation to trust in computers and humans. These specificities can be further investigated in future research by comparative experimental conditions involving human-computer, human-robot interactions and human-human interactions. In particular, given the child-like aspect of iCub, it would be interesting to compare interactions between two children vs. interaction between a child and iCub.

To continue on our findings, following the proposal of Young *et al*. (2009) that trust in functional savvy is a prerequisite for trust in social savvy, our second hypothesis was that users would not trust the social savvy uniquely. Consequently, we predicted that participants conforming to the iCub's answers in the social task would also conform in the functional task. Contrary to our hypothesis and to the current literature (Young *et al*., 2009), results show that the very limited number of participants conforming in the social task conformed less in the functional task. This result indicates that the few participants, who believed that social savvy is susceptible to be an intrinsically reliable ability of the robot, do not believe the same for functional savvy. This minority does not thus base trust in social savvy on trust in functional savvy, and on the contrary seems to assume that a social robot that has poor technical skills and little objective knowledge would not be necessary an untrustworthy social robot for this.

Globally, this first set of results leads us to conclude that, over the limited portion of participants who trusted in the robot, a restrained number considered that trust in functional savvy excludes trust in social savvy, and a even more restrained number of users who considered that trust in social savvy excludes trust in functional savvy. This mutual incompatibility between trust in functional and social savvy may represent a deeper understanding of users trust and distrust behaviors: if these behaviors have indeed been observed in previous studies with relation to the robot's level of autonomy (Heerink, 2010;

Schaefer, 2013) and adaptability (Kamide *et al.*, 2013), our study puts the accent on the nature of the task itself, since this latter concretely instantiates the double nature of the humanoid robots, which are conceived to be both functional and social devices. The result could also be symptomatic of a dichotomist view of the robot by the participants: either the robot is a machine that can assist the human in functional tasks, or it is a companion that can interact at a social level. Interestingly, while the anthropomorphic aspect of the robot could suggest a more social companion, our results indicate that the participants trusted more the robot's functional savvy. In some sense, this is contrary to a line of research in robot companions (*e.g.,* Aldebaran's Pepper, Nao, Romeo) that frequently seeks anthropomorphic shapes to enhance the robot's acceptance, while it is more likely that it could be more profitable to focus on the robot's functionalities as assistant.

With regards to the desire for control, the negative attitude towards the robots'social influence, and the simulated interaction scenario with robot, the results only partially confirm our hypothesis as well as the discussed literature.

More in detail, results concerning the influence of the desire for control on the participants' conformation indicate that, whether participants showed a high or low desire for control, they mainly tended not to conform to iCub. This confirms previous studies where users generally preferred to be in control of the robot (Syrdal *et al.*, 2007; Kamide *et al.*, 2013, Koay *et al.*, 2014; Okita *et al.*, 2012; Gilles *et al.*, 2004; Marble *et al.*, 2004). With regards to the thirteen participants conforming to iCub in the functional task, and to the three participants conforming to iCub in the social task, we might ascribe this behavior to the fact that the young age of iCub, whose appearance is inspired from the one of a 4 years old child, aroused a tolerant behavior, so that these participants did not wish to contradict iCub just like sometimes parents or teachers find it difficult to contradict a child. For example, one participant reported that she wished not to contradict the robot because it was cute, and she wanted to please it. When asked, she said the robot reminded a small child. The nice robot appearance, for some participants, was apparently not suggesting the interaction with a child, but rather with a pet.

Concerning results on the influence of the attitude towards robots, results show that the participants' trusting behaviors were independent from the negative attitude towards the social influence of the robot: whether they fear to be influenced by the robot or not, participants did not conform their decisions to those of the robot, except for the minority who conformed during functional tasks and for the even more restrained minority who conformed during social tasks. This could mean that distrust is itself a diffused symptom of a rather negative attitude towards robot, which seems to be more deeply rooted than specific fears about robots. Hence, even when the fear of robots will be overcome, still psychologists and engineers will have to deal with a generalized and substantial level of distrust amongst users.

Further analysis also showed that the more participants have an elevated desire for control (score at the DFC), the less they fear to be influenced by robots (score at the NARS-S2). These participants seem thus not to be concerned by the possibility that a robot could determine their decisions: they probably assume that they will be still in control of their tasks outcome and they would not need to rely exclusively on robots. This result seems to indicate that participants with strong desire for control interpret trust mainly as reliance – with reliance being one of the components of trust (Billing *et al*., 2012) - and that the idea of being in the condition to rely on a robot makes them unwilling to use the robot. Correlation between desire of control and negative attitudes towards robot could thus be revealing of a disposition to use the robot. Prospective developments of the present study should thus identify the participants' levels of disposition of use as a preliminary predictor of trust behavior. Moreover since this study mainly focus on robot functional and social acceptance, it shall be useful to additionally assess participants regular use of technology, by asking them if they normally use social technologies (*e.g.,* social networks, videogames, etc.), and whether they use functional technologies for social purposes (*e.g.,* using a office computer for chatting, watching a movie, etc.) as well as social technologies for functional use (*e.g.,* using social networks to retrieve useful information, selling objects, looking for a job, etc.). These can also be considered as predictors of trust behaviors.

Finally, results of the influence of the imagined scenario on the trusting behavior seem to infirm our hypothesis that imagining a collaborative interaction scenario would determine a higher level of trust, While this in contrast with the findings in the study of Kuchenbrandt *et al.* (2012) witnessing that robots acceptance increases when users imagine a collaborative scenario, this is line with the more recent study of Wullenkord *et al.* (2014), where participants who had imagined contact with a robot did not report more positive attitudes towards robots. However, in the specific case of our study, the weak influence of the imagined scenario on the participants' level of trust can be explained by the fact that the participants had difficulties in imagining the proposed interaction scenario. Apparently, participants who had been asked to imagined a competitive scenario could hardly recall that they imagined it. This can be due to the fact that while in the imagined scenario participants thought about an abstract non-specified robot, in the real interaction they were confronted with a specific robot that was child-like and with a rather sympathetic appearance. Again, the appearance of iCub could have diminished the strength of the competitive scenario and enhanced the strength of collaborative scenario in their memory. Therefore, in future research we shall vary the physical appearance of the robot to validate our hypothesis on the influence of imagined scenario on users trust.

## 6. Conclusions

This study shows that the robot acceptance is a complex dynamics characterized by a prevailing distrust on robots, and where the few trust behaviors that can be observed amongst users are significantly correlated to the nature of the task at hand. In particular, robots seem to be more easily accepted in functional than in social tasks. This is witnessed by the evidence that when confronted with tasks requiring decisions about functional issues, users trust robots more than they do when confronted with tasks requiring decision on social issues. Moreover, the minority of users who trust robots on social issues shows a significant distrust in robots on functional issues.

These results do not allow us to understand whether the observed distrust and trust behaviors depend on the fact that, despite today robots' vocation is to be at the same time a functional and social technology, users still consider robots as "socially ignorant" (Young, *et al.* 2009) or rather on the fact that trust in functional savvy and trust in social savvy are mutually exclusive because they are based upon different kinds of knowledge and skills (scientific objective knowledge and technical skills vs. subjective common-sense knowledge and adaptive skills).

However, the general distrust and the different behavior in the two tasks suggest that trust in robots cannot be assimilated to mechanisms that are typical of trust in computers neither to those that are typical of trust in humans.

Furthermore, the three observed HRI factors (desire for control, negative attitudes towards social influence of robots, and imagined interaction scenario) seem not to have influence on trust behaviors in terms of conformation of users' decisions to robot decisions.

Nonetheless, we registered a significant correlation between desire for control and negative attitudes towards robot. This correlation indicates that the more participants wish to have control on their life situations, the less they fear of being influenced by a robot, and this suggest that users with control-seeker profiles do not wish to use robots as far as they would have to rely on them. Consequently, we argued that correlation between desire for control and negative attitudes towards robots could help in quantifying the intention of use. Additionally, we proposed that the intention of use can be an interesting predictor of trust for participants whose strong desire for control implies that they reduce the complexity of trust to one of its component, which is reliance (*e.g.,* dependence from robots), and tend thus to resistance to robots rather than acceptance of them. Likewise, we have proposed that users' habits with concerns to functional and social technologies can be also investigated as predictors of trust behavior.

Finally, from a methodological point of view, our study is an attempt to establish a conceptual bond among pivotal issues that, though being recurrently used in the HRI communities such as "social and functional acceptance", "robots savvy" and "trust", still suffer of some fuzziness since they have be inherited from other domains, namely ergonomics and social psychology, and need to be adapted to social robotics in order to get a more profound understanding of users acceptance of robots.

**Acknowledgments**

This work was performed within the project EDHHI of Labex SMART (ANR-11-LABX-65) supported by French state funds managed by the ANR within the Investissements d'Avenir programme under reference ANR-11-IDEX-0004-02.
The work was partially supported by the CHART-LUTIN (EA 4004 - FED 4246).

We wish to thank Charles Ballarini for his contribution in software and experiments, Salvatore Anzalone and Joëlle Provasi for their contribution to the design of the experimental protocol. We are grateful to all the participants that took part in the experiment.

**Annex 1: The WoZ GUI**

The WoZ GUI was organized in several tabs, each dedicated to a specific task, such as controlling the robot movements - gaze, hands movements, posture (Figure 1), its speech (Figure 2), its face expressions (Figure 3), etc. The GUI events are elaborated by the actionServer module and other modules developed by the authors in previous works (Ivaldi *et al.*, 2014a, 2014b).

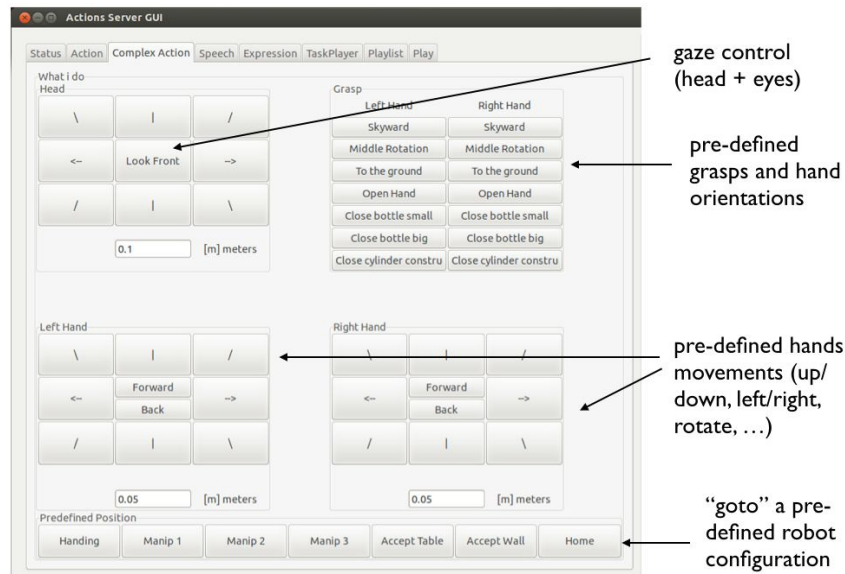

**Figure 1 -** WoZ GUI: the tab dedicated to the quick control of gaze, grasps and hands movements in the Cartesian space. The buttons send pre-defined commands to the actionsServer module, developed in (Ivaldi *et al.*, 2014). The buttons of the bottom row allow the operator to bring the robot in pre-defined postures (whole-body joint configurations): they were pre-programmed so as to simplify the control of the iCub during the experiments, in case the operator had to "bring it back" to a pre-defined configuration that could simplify the interaction for the subjects. They were useful also for prototyping and testing of the experiments.

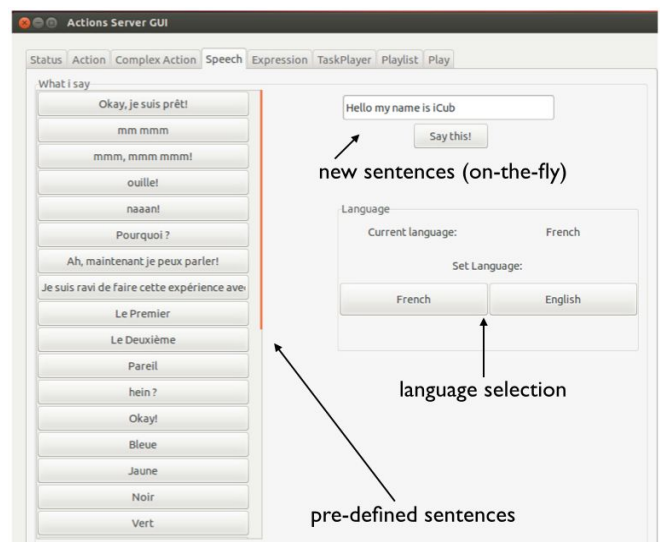

**Figure 2 -** WoZ GUI: the tab related to the robot's speech. The operator can choose between a list of pre-defined sentences

and expressions, or he can type a new sentence on-the-fly: this is done to be able to quickly formulate an answer to an unexpected request of the participant. The operator can switch between French and English speech (at the moment, the only two supported languages), even if in the experiments of this paper of course the robot was always speaking French.

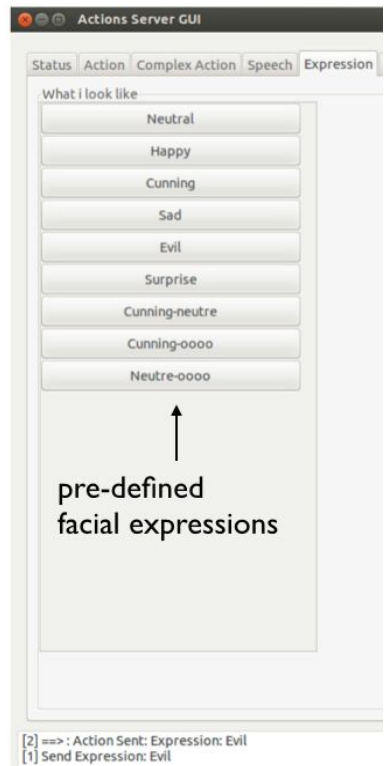

**Figure 3** – WoZ GUI: the tab related to facial expressions. The list of facial expression along with their specific realization on the iCub face (the combination of the activation of the LEDs in eyelids and mouth) is loaded from a configuration file.

# Annex 2 The NARS test and DFC test with their French translation

**Table 1:** NARS questionnaire for evaluating the negative attitude towards robots. The order of the questions follows the original questionnaire, proposed by Nomura et al. (2006). The second column reports the original questions in English. The third column reports our double translation of the questions in French.

| N. | Questionnaire Item in English | Questionnaire Item in French | Subscale |
|---|---|---|---|
| 1 | I would feel uneasy if robots really had emotions. | Je me sentirais mal à l'aise si les robots avaient réellement des émotions. | **S2** |
| 2 | Something bad might happen if robots developed into living beings. | Quelque chose de mauvais pourrait se produire si les robots devenaient des êtres vivants. | **S2** |
| 3 | I would feel relaxed talking with robots. | Je serais détendu(e) si je parlais avec des robots. | S3* |
| 4 | I would feel uneasy if I was given a job where I had to use robots. | Je me sentirais mal à l'aise dans un travail où je devrais utiliser des robots. | S1 |
| 5 | If robots had emotions, I would be able to make friends with them. | Si les robots avaient des émotions, je serai capable de devenir ami(e) avec eux. | S3 |
| 6 | I feel comforted being with robots that have emotions. | Je me sens réconforté(e) par le fait d'être avec des robots qui ont des émotions. | S3* |
| 7 | The word "robot" means nothing to me. | Le mot ''robot'' ne signifie rien pour moi. | S1 |
| 8 | I would feel nervous operating a robot in front of other people. | Je me sentirais nerveux/nerveuse de manœuvrer un robot devant d'autres personnes. | S1 |
| 9 | I would hate the idea that robots or artificial intelligences were making judgments about things. | Je détesterais que les robots ou les intelligences artificielles fassent des jugements sur des choses. | S1 |
| 10 | I would feel very nervous just standing in front of a robot. | Le simple fait de me tenir face à un robot me rendrait très nerveux/nerveuse. | S1 |
| 11 | I feel that if I depend on robots too much, something bad might happen. | Je pense que si je dépendais trop fortement des robots, quelque chose de mauvais pourrait arriver. | **S2** |
| 12 | I would feel paranoid talking with a robot. | Je me sentirais paranoïaque de parler avec un robot. | S1 |
| 13 | I am concerned that robots would be a bad influence on children. | Je suis préoccupé(e) par le fait que les robots puissent avoir une mauvaise influence sur les enfants. | **S2** |
| 14 | I feel that in the future society will be dominated by robots. | Je pense que dans le futur la société sera dominée par les robots. | **S2** |
| | ● = reverse item  ● Only S2 items where used in the present study | | |

**Table 2:** Desire For Control questionnaire. The order of the questions follows the original questionnaire, proposed by Burger and Cooper (1979). The second column reports the original questions in English. The third column reports our double translation of the questions in French.

| N. | Questionnaire Item in English | Questionnaire Item in French |
|---|---|---|
| 1 | I prefer a job where I have a lot of control over what I do and when I do it. | Je préfère un travail où j'ai un contrôle important sur ce que je fais et quand je le fais. |
| 2 | I enjoy political participation because I want to have as much of a say in running government as possible. | J'apprécie la participation politique parce que je veux avoir, autant que possible, la possibilité de m'exprimer dans le fonctionnement d'un gouvernement. |
| 3 | I try to avoid situations where someone else tells me what to do. | J'essaie d'éviter les situations dans lesquelles quelqu'un me dit ce que je dois faire. |
| 4 | I would prefer to be a leader than a follower. | Je préfère être un leader plutôt qu'un suiveur. |
| 5 | I enjoy being able to influence the actions of others. | J'apprécie de pouvoir influencer les actions des autres. |
| 6 | I am careful to check everything on an automobile before I leave for a long trip. | Je fais attention à tout vérifier dans une voiture avant de partir pour un long voyage. |
| 7* | Others usually know what is best for me. | Les autres savent généralement ce qui est bon pour moi. |
| 8 | I enjoy making my own decisions. | J'apprécie de pouvoir prendre mes propres décisions. |
| 9 | I enjoy having control over my own destiny. | J'apprécie d'avoir le contrôle sur ma propre destinée. |
| 10* | I would rather someone else took over the leadership role when I'm involved in a group project. | Je préfère que quelqu'un d'autre prenne le rôle de leader quand je suis impliqué(e) dans un projet de groupe. |
| 11 | I consider myself to be generally more capable of handling situations than others are. | Je me considère comme généralement plus capable de gérer les situations que les autres. |
| 12 | I'd rather run my own business and make my own mistakes than listen to someone else's orders. | Je préfère diriger ma propre affaire et faire mes propres erreurs qu'écouter les ordres de quelqu'un d'autre. |
| 13 | I like to get a good idea of what a job is all about before I begin. | J'aime avoir une idée globale et claire d'un travail avant de le commencer. |
| 14 | When I see a problem I prefer to do something about it rather than sit by and let it continue. | Quand je rencontre un problème, je préfère faire quelque chose à propos de celui-ci plutôt que de rester passif. |
| 15 | When it comes to orders, I would rather give them than receive them. | En ce qui concerne les ordres, je préfère en donner qu'en recevoir. |
| 16* | I wish I could push many of life's daily decisions off on someone else. | J'aimerais pouvoir me décharger du poids des décisions du quotidien sur quelqu'un d'autre. |
| 17 | When driving, I try to avoid putting myself in a situation where I could be hurt by someone else's mistake. | Sur la route, j'essaie d'éviter de me mettre dans des situations où je pourrais être blessé à cause de quelqu'un d'autre. |
| 18 | I prefer to avoid situations where someone else has to tell me what is I should be doing. | Je préfère éviter les situations dans lesquelles quelqu'un doit me dire ce que je devrais faire. |
| 19* | There are many situations in which I would prefer only one choice rather than having to make a decision. | Dans de nombreuses situations, je préfère n'avoir qu'une seule option plutôt que de devoir faire un choix entre plusieurs options. |
| 20* | I like to wait and see if someone else is going to solve a problem so that I don't have to be bothered by it. | J'aime attendre de voir si quelqu'un d'autre va résoudre un problème de sorte que je n'ai pas à m'en soucier. |
| * = reverse item | | |

# Annex 3: The 3 sub-tasks designed to test the trust in functional savvy

**Evaluating sounds:**
The experimenter explained that the goal of the evaluation was to determine the most acute sounds between two. The experimenter would use the computer on her right to play the two sound stimuli consecutively, saying "first" and "second" before each sound (in French, "le premier" and "le deuxième"). The sound stimuli were sinusoidal waveforms at a pure frequency, lasting 1.5 seconds, and were generated by the Audacity software.
The frequencies of the sounds were:
1. ("equal" or "50-50" case): 450 Hz and 450 Hz;
2. ("different" case): 117Hz and 200Hz;
3. ("ambiguous" case): 450Hz and 455Hz;
4. ("ambiguous" case): 100Hz and 110Hz.

For each question, the experimenter asked "Which is the most acute sound: the first or the second?". We remark that for most of the participants, this was the most difficult task as the environmental conditions were probably not optimal for evaluating the sounds (background noise from computers etc.).

**Evaluating images:**
The experimenter explained that the goal of this evaluation was to determine the dominant color in an image. Three colors were suggested for each image. The experimenter would use the computer on her right to show the images consecutively.
The images are shown in Figure.
For each image, the suggested colors were:
1. ("equal" or "50-50" case): abstract pattern; possible colors: blue, yellow, black. Here the robot always answered blue or yellow, depending on the participant's answer.
2. ("different" case): a red ladybird on a green leaf; possible colors: green, yellow, red. Here the robot always choose "green" as the most dominant color.
3. ("ambiguous" case): industrial scenario; possible colors: green, blue, grey. Here the robot always answered green or blue, depending on the participant's answer.
4. ("ambiguous" case): grey worm on a green leaf; possible colors: green, grey, black. Here the robot always answered green or grey, depending on the participant's answer.

For each image, the experimenter asked "Which is the dominant color in this image: the <first color>, the <second color> or the <third color>?"

**Evaluating weights:**
The experimenter explained that the goal of this evaluation was to determine the heavier object between two. Four pairs of bottles, with different or similar shape, color and weight (filled or not) were used. The bottles are shown in Figure REF.
For each pair, the weights are:
1. ("equal" or "50-50" case): two white bottles filled with big beans, identical in appearance and weight (42g);
2. ("different" case): two bottles filled with big beans, with different appearance and different weight (41g and 73g);
3. ("ambiguous" case): two bottles filled with small beans, with similar appearance and slightly different weight (50g and 63g);
4. ("ambiguous" case): two bottles filled with big beans, with different appearance and slightly different weight (65g and 57g).

For each pair of bottles, the experimenter asked "Which bottle is heaviest: the first or the second?". The experimenter always took the bottle with number one in her left hand, and number two in her right hand. She gave the bottles to the participant, who could take his/her time to evaluate, sometimes swapping the bottles in the hands. Then the experimenter gave the bottles to the robot. The operator would activate a pre-designed arm trajectory, opening the hands. The experimenter put the bottles in the middle of the palm, saying "Close the hands"; the robot would then grasp the bottles with a pre-designed grasp (see Figure). The experimenter repeated the question. Given the different sizes of the bottles, two different pre-programmed grasps were used. When the robot had said "First" or "Second", the experimenter said "Open the hands", retrieved the bottles and gave them back to the participant for getting his/her final decision.